\documentclass{article}
\usepackage[dblblindworkshop, final]{neurips_2026}
\usepackage{graphicx} 
\usepackage[font=small,labelfont=bf]{caption}
\usepackage{subfigure}
\usepackage[utf8]{inputenc} 
\usepackage[T1]{fontenc}    
\usepackage{hyperref}       
\usepackage{url}            
\usepackage{booktabs}       
\usepackage{amsfonts}       
\usepackage{amssymb}
\usepackage{amsmath}
\usepackage{nicefrac}       
\usepackage{microtype}      
\usepackage{xcolor}         
\usepackage{longtable}
\usepackage{amsthm}
\newtheorem{theorem}{Theorem}
\newtheorem*{theorem*}{Theorem}
\newtheorem{corollary}{Corollary}
\newtheorem{definition}{Definition}

\workshoptitle{AI for Stochastic Dynamics}
 
\title{Stochastic Dynamics of Long-Horizon Generative Models}

\author{%
  András Horváth \\
  Faculty of Information Technology and Bionics\\
  Peter Pazmany Catholic University\\
  Budapest, Hungary\\
  \texttt{horvath.andras@itk.ppke.hu} \\
}

\date{March 2026}

\begin{document}

\maketitle

\begin{abstract}
Long-horizon generation in autoregressive and iterative generative models can exhibit repetition, reduced contextual sensitivity, and loss of diversity. We study these behaviors from a dynamical perspective by analyzing how separation between generation trajectories evolves in representation space. Using an approximate-contraction condition, in which expected pairwise distance contracts by a factor $\lambda$ up to an additive stochastic scale $\epsilon$, we define an effective relaxation timescale $\tau=1/(1-\lambda)$ and long-horizon separation scale $r=\epsilon/(1-\lambda)$.

We evaluate this framework across long-horizon autoregressive language generation and iterative image outpainting. Across models and generation settings, pairwise representation distances initially decrease and subsequently fluctuate around a nonzero long-horizon scale. In language generation, the effective parameters vary systematically with decoding temperature and model architecture: higher temperatures generally produce larger stochastic scales and, in several models, stronger fitted contraction. In image outpainting, $\lambda$ and $\epsilon$ vary substantially with generation model and extension length, while the resulting separation scale remains comparatively stable, with values near $1.3$ across most configurations. Because representations are unit-normalized, these separation scales correspond to mean pairwise cosine similarities and partly reflect static representation geometry such as anisotropy. Noise-reduced Lyapunov-like diagnostics provide a complementary measure of local trajectory sensitivity.

The framework provides an empirical description of long-horizon representation dynamics across generative systems, without requiring strict attractors or implying that representation-space contraction causes output-level degeneration.
\end{abstract}

\section{Introduction}

Generative models are trained through local prediction objectives, but their behavior over hundreds or thousands of generation steps is inherently dynamical. In autoregressive generation, each generated state becomes part of the input to the next step, allowing stochastic effects and state-dependent transformations to accumulate. At long horizons, this can lead to repetition, reduced contextual sensitivity, and loss of diversity~\cite{holtzman2019curious,wang2023shall,naparstek2026token}. Understanding such behavior therefore requires studying not only individual predictions, but also the geometry and stability of the resulting trajectories.

We view this process through the evolution of representation-space distances. Given two trajectories initialized from different states, we ask how their expected separation changes with the generation horizon. To characterize this evolution, we use an effective stochastic description that separates the deterministic component of the dynamics from stochastic effects arising from sampling and generation. This description is intended as an effective model of the observed dynamics rather than a mechanistic decomposition of the underlying neural network.

Rather than assuming global contraction of the underlying model, we characterize contraction empirically through expected pairwise distances. Specifically, we consider an approximate contraction condition in which the expected separation at the next step is bounded by a fraction $\lambda<1$ of the current separation plus an additive stochastic scale $\epsilon$. This yields a geometric bound with two components: an exponentially decaying dependence on the initial separation and a nonzero long-horizon scale determined by the balance between contraction and stochasticity.

Our goal is to provide an empirical dynamical characterization rather than a causal explanation of degeneration. In particular, contraction in representation space does not imply convergence to a unique state, semantic equivalence of outputs, or a causal mechanism for repetition. Instead, we use the contraction--stochasticity decomposition as a quantitative description of long-horizon trajectory geometry.

\paragraph{Related work.}
Prior work has studied degeneration and quality--diversity trade-offs in neural text generation~\cite{holtzman2019curious,fan2018hierarchical,welleck2019neural,wiher2022decoding}, as well as underutilization of distant context in transformer models~\cite{khandelwal2018sharp,liu2024lost,press2021train}. Repetition has been linked to the self-reinforcement of repeated $n$-grams~\cite{xu2022loop}, to the anisotropy of contextual representations~\cite{ethayarajh2019contextual}, and to attention sinks~\cite{xiao2024sinks}. Dynamical-systems perspectives have been applied to neural-network stability, representation learning, and autoregressive forecasting~\cite{bai2019deep,chang2017multi,haber2018stable,gomez2017reversible,chen2018neural,zhang2026closing}, with recent work also considering control-theoretic views of autoregressive representations~\cite{kong2024representation,naparstek2026token}. Closest to our motivation, Du and Tanaka-Ishii~\cite{du2026escaping} interpret mode collapse as a geometric collapse of the generation trajectory in representation space and regulate it during decoding. We focus instead on a simple empirical description of long-horizon trajectory contraction and stochasticity that can be applied uniformly across generative systems.

Our main contributions are:
\begin{itemize}
\item We formulate long-horizon generation as an effective stochastic process in representation space and adopt an approximate-contraction criterion based on expected pairwise distance, building on established stochastic contraction results~\cite{pham2009contraction,diaconis1999iterated}.
\item We use the resulting geometric distance bound to define an empirical relaxation timescale and long-horizon separation scale, together with a simple estimation procedure.
\item We characterize these quantities across autoregressive language models and iterative image-generation systems, identifying systematic dependence on decoding temperature, architecture, and outpainting extent, and relating the separation scale to static representation geometry.
\end{itemize}

\section{Theoretical Framework}

The recurrence used here is a standard consequence of contraction with bounded stochastic perturbations and is not new. Definition~\ref{def:contraction} corresponds to the stochastic contraction condition of Pham, Tabareau and Slotine~\cite[Def.~1]{pham2009contraction}, with the same two parameters. The resulting bound (Theorem~\ref{thm:distance}) is~\cite[Thm.~2, Eq.~2.10]{pham2009contraction}, and its geometric-series form appears in the analysis of iterated random functions by Diaconis and Freedman~\cite[Eq.~2.4, Sec.~5]{diaconis1999iterated}, who also trace it to earlier work. The asymptotic bound~\eqref{eq:asymptotic_bound} restates~\cite[Eq.~2.13]{pham2009contraction} and~\cite[Thm.~4.1]{zhang2026closing}. We include the derivation only for completeness and to establish notation. Our contribution is the use of this bound as an empirical characterization of long-horizon generative trajectories, together with the associated estimation procedure (Sec.~\ref{sec:estimation}).

\subsection{Effective Stochastic Dynamics}

Let $x_t$ denote the generated state at step $t$ (the generated prefix for language models, the current image for outpainting) and $z_t=\phi(x_t)\in\mathbb{R}^d$ its representation under an embedding function $\phi$. We use the effective stochastic update
\begin{equation}
z_{t+1}=G(z_t)+\xi_t,
\qquad
\mathbb{E}[|\xi_t|]\leq\epsilon,
\label{eq:dynamics}
\end{equation}
where $G$ captures the effective deterministic dynamics, $\xi_t$ represents stochastic effects, and $\epsilon\geq0$ is an effective stochastic scale. The decomposition is not intended to identify an explicit component of the neural network.

\subsection{Approximate Contraction and Long-Horizon Separation}

Let $d_t=\mathbb{E}[|z_t-z_t'|]$ denote the expected distance between two trajectories. We characterize contraction-like trajectory dynamics through the following condition.

\begin{definition}[Approximate contraction; cf.~\cite{pham2009contraction}, Def.~1]
\label{def:contraction}
A stochastic representation process satisfies a $(\lambda,\epsilon)$ approximate-contraction condition if
\begin{equation}
\mathbb{E}[|z_{t+1}-z_{t+1}'|]
\leq
\lambda\mathbb{E}[|z_t-z_t'|]+\epsilon,
\label{eq:contraction}
\end{equation}
where $\lambda\in[0,1)$ and $\epsilon\geq0$.
\end{definition}

Here $\lambda$ describes the effective reduction in trajectory separation within the fitted recurrence, while $\epsilon$ captures an additive stochastic scale. Importantly, $\lambda$ is an empirical average-case quantity and is not a worst-case Lipschitz constant or a model-intrinsic contraction coefficient.

\begin{theorem}[Expected distance bound;~\cite{pham2009contraction}, Thm.~2;~\cite{diaconis1999iterated}]
\label{thm:distance}
Under Definition~\ref{def:contraction},
\begin{equation}
\mathbb{E}[|z_t-z_t'|]
\leq
\lambda^t\mathbb{E}[|z_0-z_0'|]
+
\frac{\epsilon}{1-\lambda}(1-\lambda^t).
\label{eq:distance_bound}
\end{equation}
\end{theorem}

The result follows by recursively applying Eq.~\eqref{eq:contraction}. The first term captures the dependence on the initial separation and decays geometrically for $\lambda<1$; the second is the accumulated stochastic contribution. We therefore define
\begin{equation}
\tau=\frac{1}{1-\lambda},
\qquad
r=\frac{\epsilon}{1-\lambda},
\label{eq:timescale_radius}
\end{equation}
where $\tau$ is an effective relaxation timescale and $r$ is the recurrence-implied long-horizon separation scale. In the long-horizon limit (cf.~\cite[Eq.~2.13]{pham2009contraction},~\cite[Thm.~4.1]{zhang2026closing}),
\begin{equation}
\limsup_{t\rightarrow\infty}
\mathbb{E}[|z_t-z_t'|]\leq r.
\label{eq:asymptotic_bound}
\end{equation}

Thus, under the approximate-contraction condition, the framework gives an upper bound on expected long-horizon separation without requiring convergence to a unique representation.

\subsection{Estimation of Effective Dynamical Parameters}
\label{sec:estimation}

For trajectory pairs $i=1,\dots,N$ with distances $d_t^{(i)}=|z_t^{(i)}-z_t^{\prime(i)}|$ and mean $\bar d_t=\frac{1}{N}\sum_{i}d_t^{(i)}$, we fit
\begin{equation}
\bar d_{t+1}\approx\lambda\bar d_t+\epsilon,
\label{eq:empirical_fit}
\end{equation}
and compute $\tau$ and $r$ from Eq.~\eqref{eq:timescale_radius}.

These fitted quantities are descriptive statistics of the observed distance dynamics. They are not assumed to be exact, time-independent properties of the underlying neural system, and may depend on the representation, initialization distribution, sampling procedure, and fitting interval. An unconstrained least-squares fit can return a slightly negative slope in flat, noise-dominated regimes; such values do not satisfy Definition~\ref{def:contraction} and are interpreted as $\lambda\approx0$, $r\approx\epsilon$.

\paragraph{Relation to plateau statistics.}
Because the regression~\eqref{eq:empirical_fit} is fitted over the full horizon, which is dominated by the plateau, the fitted parameters are strongly constrained by plateau statistics. For ordinary least squares, $\hat\epsilon=\bar y-\hat\lambda\bar x$, where $\bar x,\bar y$ are the means of $\bar d_t$ and $\bar d_{t+1}$. When $\bar x\approx\bar y\approx d^\ast$, this gives $\hat r=\hat\epsilon/(1-\hat\lambda)\approx d^\ast$: $\hat r$ essentially reproduces the mean plateau distance, and $\hat\lambda$ is closer to the lag-one autocorrelation of plateau fluctuations than to the rate of the initial transient. The agreement between predicted and observed long-horizon separation (Fig.~\ref{fig:imgperturbation}, right) is therefore expected by construction and should be read as a consistency check, not as independent validation. Likewise, since $\Lambda_{\mathrm{NR}}$ (Sec.~\ref{sec:lyap}) averages $\log((d_{t+1}-\hat\epsilon)/d_t)$ and $d_{t+1}-\hat\epsilon\approx\hat\lambda d_t$ under the fit, part of the agreement between $\Lambda_{\mathrm{NR}}$ and $\log\hat\lambda$ (Fig.~\ref{fig:ljapunov}a) is implied by the estimator. Finally, $\hat\epsilon>0$ is obtained even for greedy decoding ($T=0$), because pairs are drawn from different prompts. $\epsilon$ therefore captures persistent inter-trajectory differences as well as sampling noise, and is best read as an additive separation floor rather than as pure sampling noise.

\subsection{Local Sensitivity Diagnostic}
\label{sec:lyap}

To complement the global distance recurrence, we consider a finite-time Lyapunov-like diagnostic $\Lambda_T=\frac{1}{T}\log(|z_T-z_T'|/|z_0-z_0'|)$. Because stochastic sampling can dominate short-horizon separation, we use a noise-reduced version, denoted $\Lambda_{\mathrm{NR}}$: a finite-time Lyapunov-like exponent in which the fitted additive term $\epsilon$ is subtracted from each step's distance before computing growth rates (Appendix~B.2).

We interpret $\Lambda_{\mathrm{NR}}$ only as an empirical local-sensitivity diagnostic. It is not a rigorous stochastic Lyapunov exponent. Comparing it with $\log\lambda$ provides a consistency check between local perturbation sensitivity and the global fitted distance dynamics; agreement is partly induced by the estimator (Sec.~\ref{sec:estimation}) and should not be read as independent confirmation.

\section{Experiments}
All codes are available at \url{https://github.com/horan85/LongHorizonContraction} for the sake of reproducability.

\subsection{Text-Based Experiments}

\paragraph{Setup.}
We use 1{,}000 diverse prompts drawn from Wikipedia, news, and LLM-generated text. For each prompt, we generate $T=1000$ tokens across nine autoregressive language models (Table~\ref{tab:full}) and nine decoding temperatures $T\in\{0.0,0.1,0.2,0.3,0.4,0.7,1.0,1.2,1.5\}$ ($T=0$: greedy decoding; otherwise temperature scaling followed by top-$k$ sampling with $k=40$) and disabling EOS so that all trajectories have the same horizon. Penultimate-layer representations are extracted every 10 generation steps and $\ell_2$-normalized. Pairwise distances are measured primarily using Euclidean distance, with cosine distance used as a qualitative comparison.

\begin{figure}[t]
\centering
\includegraphics[width=0.45\linewidth]{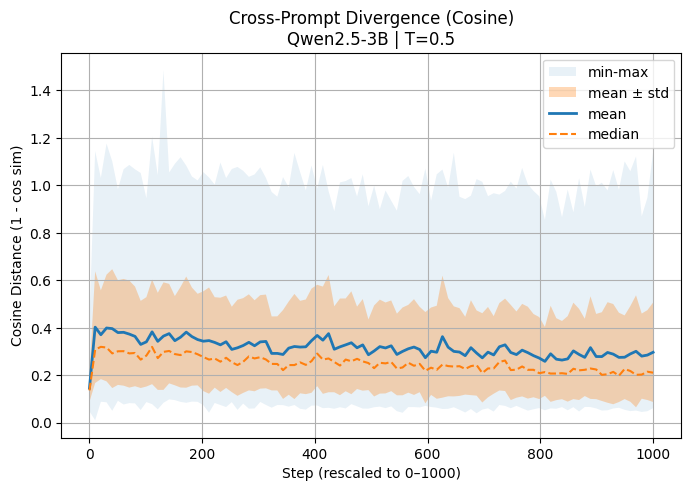}
\includegraphics[width=0.45\linewidth]{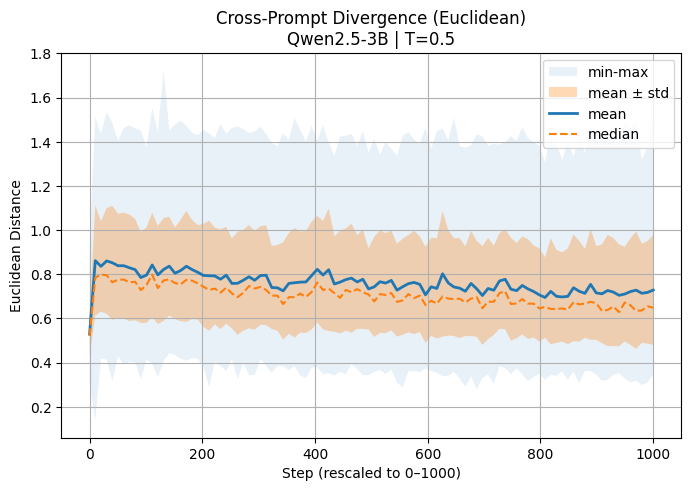}
\caption{
Evolution of cosine and Euclidean distances between randomly selected generation trajectories for Qwen2.5-3B at temperature $0.5$. Pairwise representation distances decrease over the generation horizon and subsequently fluctuate around a nonzero scale.
}
\label{fig:perturbation}
\end{figure}

\paragraph{Trajectory separation.}
Figure~\ref{fig:perturbation} shows representative trajectories for Qwen2.5-3B at temperature $0.5$. Pairwise representation distances decrease substantially during generation and subsequently fluctuate around a nonzero long-horizon scale. This behavior is consistent with the proposed contraction--stochasticity description, although it does not imply convergence of the generated text or a unique attractor.

\paragraph{Effective dynamics.}
We estimate $\lambda$ and $\epsilon$ by fitting $d_{t+1}\approx\lambda d_t+\epsilon$, and obtain $\tau$ and $r$ from Eq.~\eqref{eq:timescale_radius}. $(\lambda,\epsilon)$ are fitted on distances between 1{,}000 randomly sampled pairs of baseline trajectories from different prompts, whereas $\Lambda_{\mathrm{NR}}$ is computed from baseline--perturbed pairs (see \emph{Local sensitivity} below). These parameters are treated as descriptive statistics of each generation setting rather than model-intrinsic constants.

\begin{figure}[t]
\centering
\begin{minipage}{0.3\linewidth}
\includegraphics[width=\linewidth]{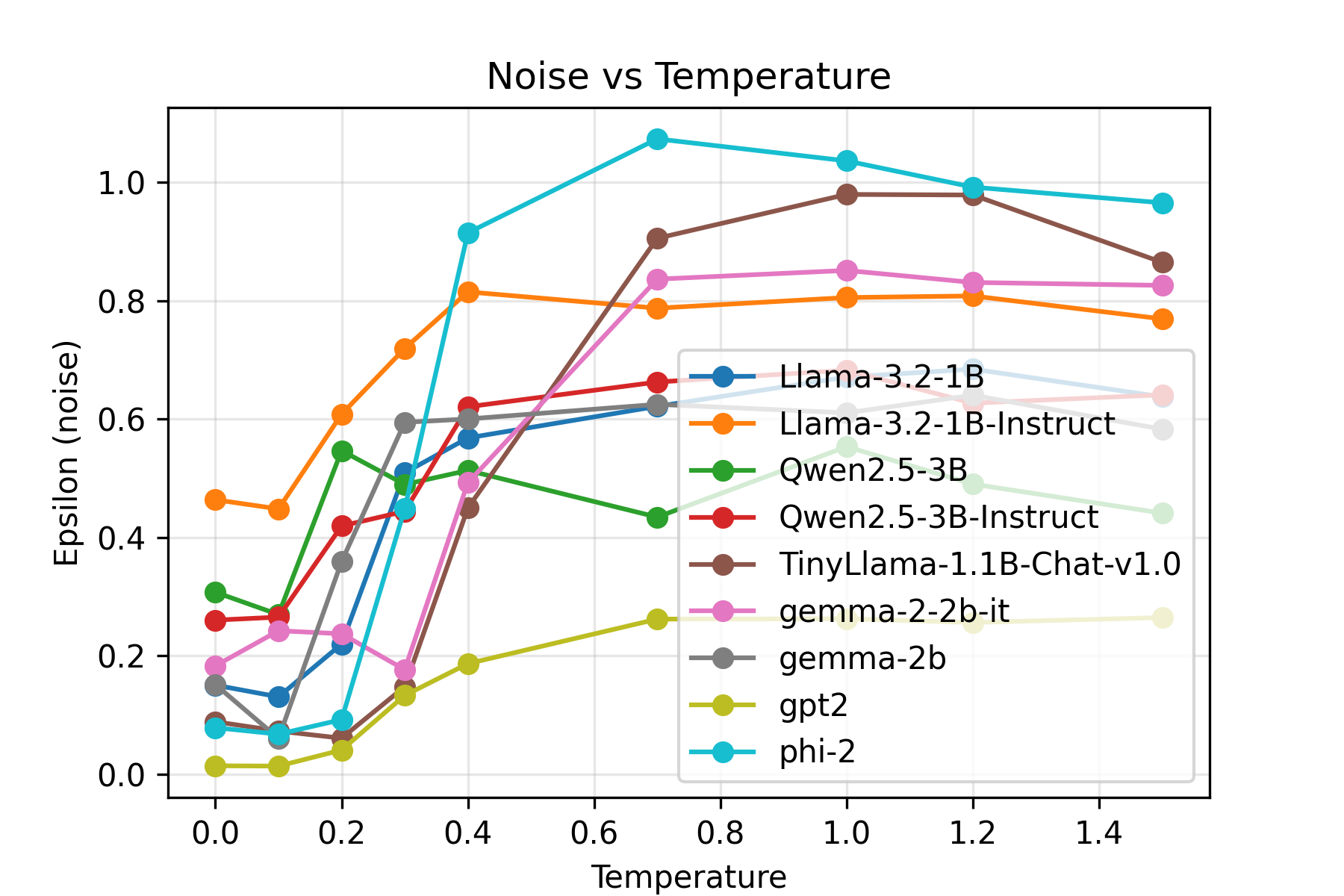}
\centerline{(a)}
\end{minipage}\hfill
\begin{minipage}{0.3\linewidth}
\includegraphics[width=\linewidth]{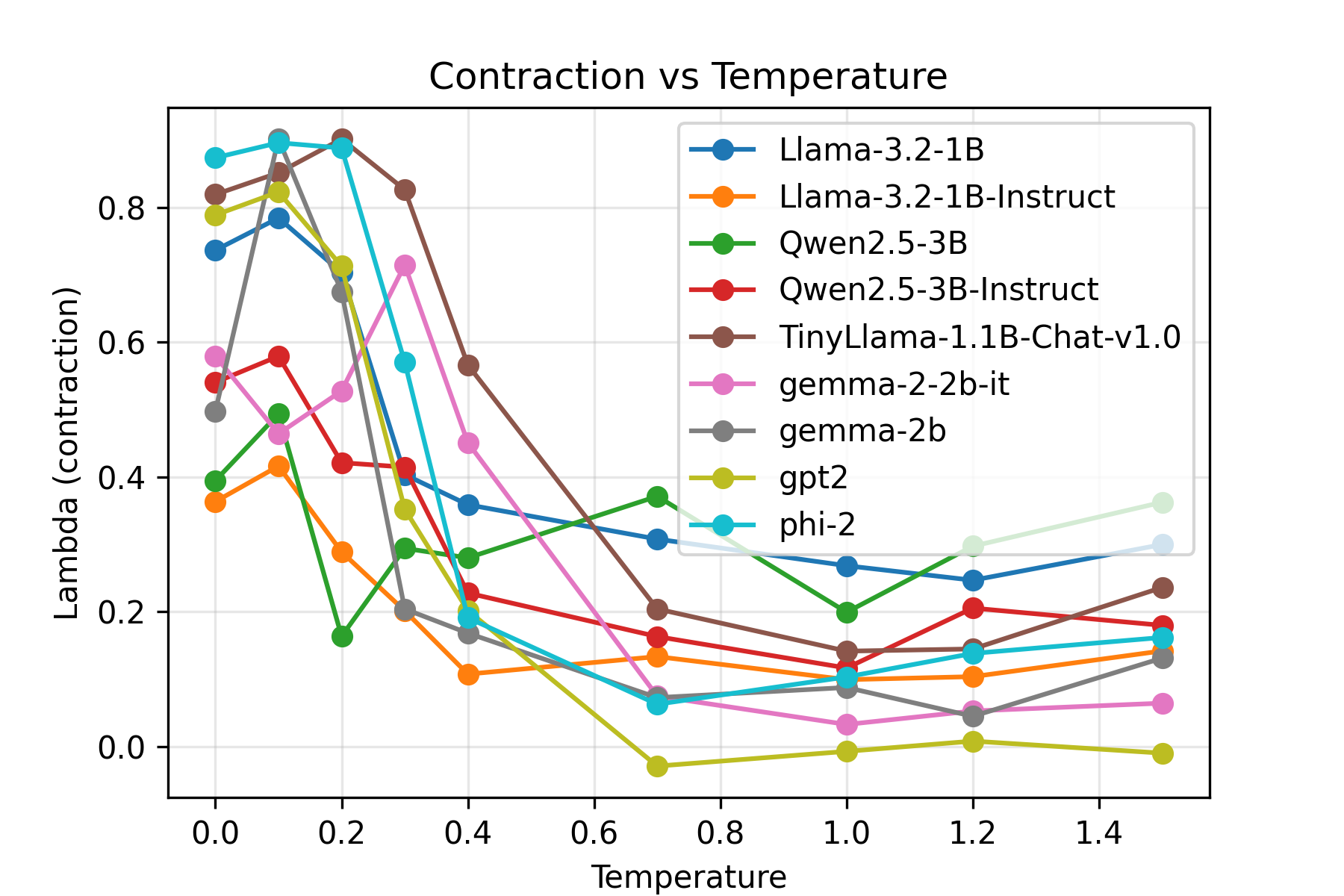}
\centerline{(b)}
\end{minipage}\hfill
\begin{minipage}{0.22\linewidth}
\includegraphics[width=\linewidth]{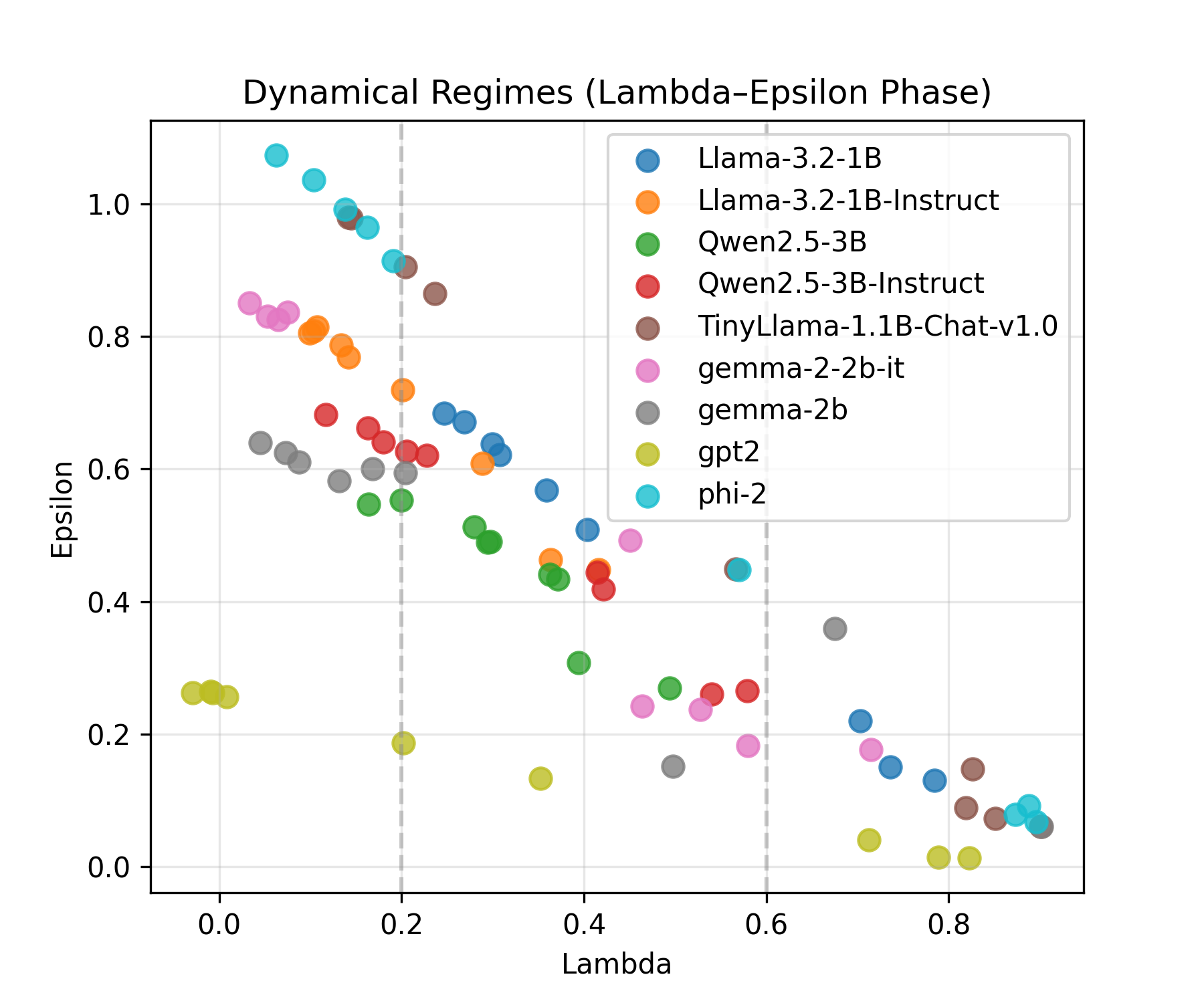}
\centerline{(c)}
\end{minipage}
\caption{
Effective dynamical parameters across models and temperature regimes.
(a,b) Estimated stochastic scale $\epsilon$ and contraction coefficient $\lambda$ as functions of temperature.
(c) Relationship between $\lambda$ and $\epsilon$ across evaluated settings.
}
\label{fig:epsilonphase}
\end{figure}

\paragraph{Temperature dependence.}
Figure~\ref{fig:epsilonphase}a,b shows the dependence of $\epsilon$ and $\lambda$ on decoding temperature. Increasing temperature generally increases $\epsilon$, while $\lambda$ tends to decrease in several models. Thus, higher temperatures are associated with stronger stochastic variability and faster fitted decay of trajectory separation. The magnitude of these effects varies substantially across architectures, as shown in Table~\ref{tab:main}.

The resulting separation scale $r$ need not decrease when $\lambda$ decreases, because changes in $\epsilon$ can compensate for changes in contraction strength.

\begin{figure}[t]
\centering
\begin{minipage}{0.35\linewidth}
\includegraphics[width=\linewidth]{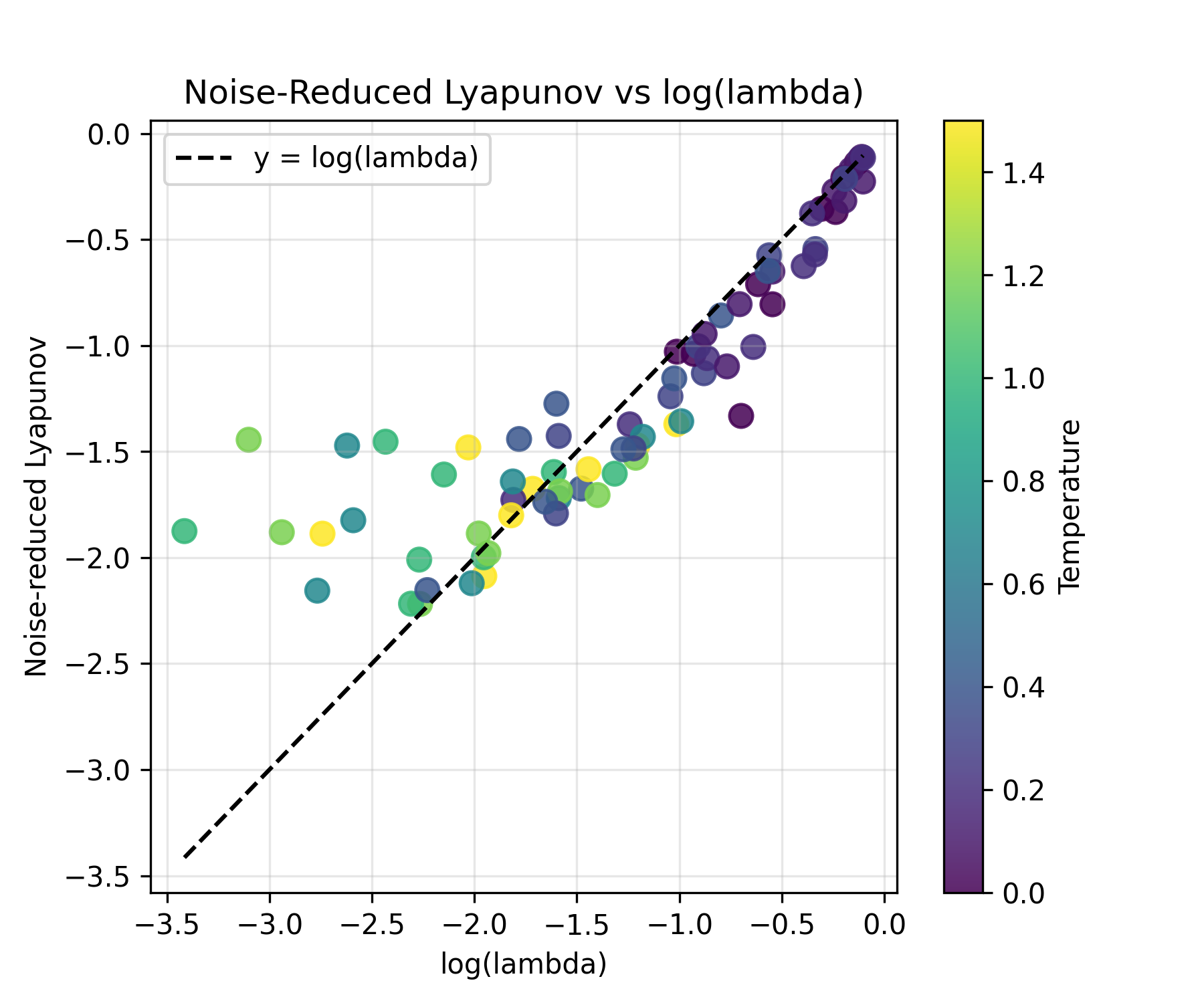}
\centerline{(a)}
\end{minipage}\hfill
\begin{minipage}{0.4\linewidth}
\includegraphics[width=\linewidth]{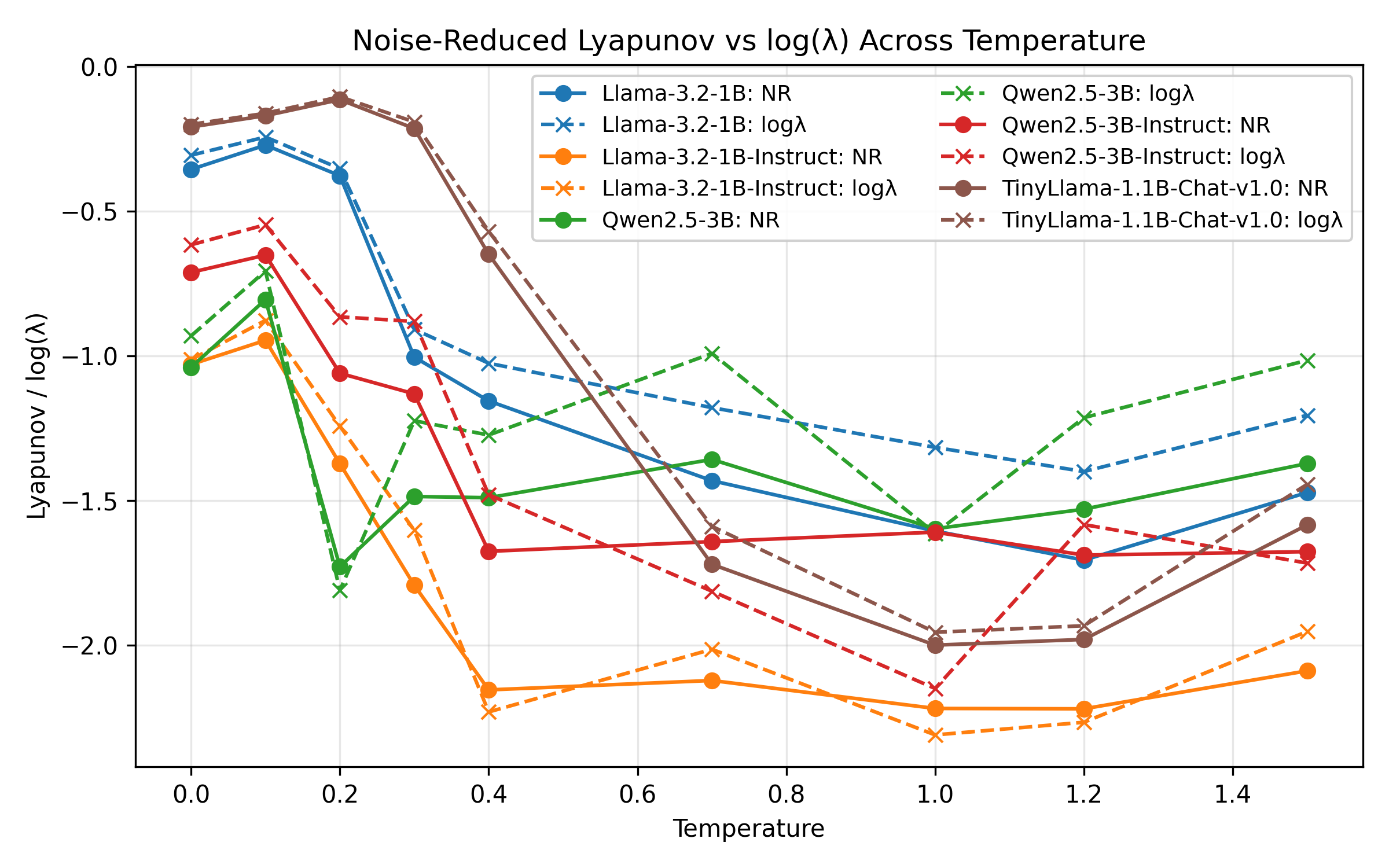}
\centerline{(b)}
\end{minipage}\hfill
\caption{
Noise-reduced Lyapunov-like analysis across language models and temperatures.
(a) Comparison of $\Lambda_{\mathrm{NR}}$ and $\log(\lambda)$.
(b) Dependence of the two stability measures on temperature. The comparison provides an empirical consistency check between local and global trajectory stability; part of the agreement is implied by the estimator (Sec.~\ref{sec:estimation}).
}
\label{fig:ljapunov}
\end{figure}

\paragraph{Local sensitivity.}
We additionally perturb the initial conditions by randomly swapping two input tokens and generate 10 perturbed trajectories for each of the 1{,}000 baseline trajectories. Figure~\ref{fig:ljapunov} compares the resulting noise-reduced Lyapunov-like diagnostic $\Lambda_{\mathrm{NR}}$ with $\log\lambda$. The comparison provides an empirical consistency check between local perturbation sensitivity and global trajectory contraction.

\paragraph{Dynamical regimes.}
Figure~\ref{fig:epsilonphase} summarizes the effective parameter space. Across the evaluated settings, larger stochastic scales generally coincide with smaller fitted values of $\lambda$. Low-temperature configurations tend to occupy a low-$\epsilon$, higher-$\lambda$ regime, whereas higher-temperature configurations exhibit larger $\epsilon$ and smaller $\lambda$. For some models (Qwen2.5-3B, Gemma-2B) these opposing changes leave $r$ nearly constant across temperature. For others (GPT-2, Llama-3.2-1B, TinyLlama, Phi-2), $r$ increases from low to intermediate temperatures and saturates for $T\geq0.4$. This illustrates why contraction and stochasticity should be considered jointly.

\begin{table}[t]
\centering
\footnotesize
\setlength{\tabcolsep}{4pt}
\begin{tabular}{llccccc}
\toprule
Model & Temp regime & $\lambda$ & $\epsilon$ & $r$ & $\Lambda_{\mathrm{NR}}$ & $\bar c \approx 1-r^2/2$\\
\midrule
GPT-2 & Low (0.0--0.3)  & $0.67\pm0.22$ & $0.05\pm0.06$ & $0.12\pm0.06$ & $-0.62\pm0.42$ & 0.99\\
      & Mid (0.4--0.7)  & $0.09\pm0.16$ & $0.22\pm0.05$ & $0.24\pm0.01$ & $-1.14\pm0.18$ & 0.97\\
      & High (1.0--1.5) & $0.00\pm0.01^{\dagger}$ & $0.26\pm0.00$ & $0.26\pm0.00$ & $-1.01\pm0.03$ & 0.97\\
\midrule
Qwen2.5-3B & Low (0.0--0.3) & $0.34\pm0.14$ & $0.40\pm0.14$ & $0.60\pm0.09$ & $-1.26\pm0.42$ & 0.82\\
           & Mid (0.4--0.7) & $0.33\pm0.06$ & $0.47\pm0.06$ & $0.70\pm0.02$ & $-1.42\pm0.09$ & 0.75\\
           & High (1.0--1.5)& $0.29\pm0.08$ & $0.49\pm0.06$ & $0.69\pm0.00$ & $-1.50\pm0.12$ & 0.76\\
\midrule
Qwen2.5-3B-Instruct & Low (0.0--0.3) & $0.49\pm0.08$ & $0.35\pm0.10$ & $0.67\pm0.09$ & $-0.89\pm0.24$ & 0.78\\
                    & Mid (0.4--0.7) & $0.20\pm0.05$ & $0.64\pm0.03$ & $0.80\pm0.01$ & $-1.66\pm0.02$ & 0.68\\
                    & High (1.0--1.5)& $0.17\pm0.05$ & $0.65\pm0.03$ & $0.78\pm0.01$ & $-1.66\pm0.04$ & 0.69\\
\midrule
Gemma-2B & Low (0.0--0.3) & $0.57\pm0.29$ & $0.29\pm0.24$ & $0.69\pm0.33$ & $-0.90\pm0.58$ & 0.76\\
         & Mid (0.4--0.7) & $0.12\pm0.07$ & $0.61\pm0.02$ & $0.70\pm0.03$ & $-1.46\pm0.02$ & 0.76\\
         & High (1.0--1.5)& $0.09\pm0.04$ & $0.61\pm0.03$ & $0.67\pm0.00$ & $-1.46\pm0.02$ & 0.78\\
\midrule
Llama-3.2-1B & Low (0.0--0.3) & $0.66\pm0.17$ & $0.25\pm0.18$ & $0.69\pm0.13$ & $-0.50\pm0.34$ & 0.76\\
             & Mid (0.4--0.7) & $0.33\pm0.04$ & $0.59\pm0.04$ & $0.89\pm0.01$ & $-1.29\pm0.19$ & 0.60\\
             & High (1.0--1.5)& $0.27\pm0.03$ & $0.66\pm0.02$ & $0.91\pm0.00$ & $-1.59\pm0.12$ & 0.58\\
\midrule
Phi-2 & Low (0.0--0.3) & $0.81\pm0.16$ & $0.17\pm0.18$ & $0.78\pm0.19$ & $-0.24\pm0.23$ & 0.69\\
      & Mid (0.4--0.7) & $0.13\pm0.09$ & $0.99\pm0.11$ & $1.14\pm0.01$ & $-1.95\pm0.30$ & 0.35\\
      & High (1.0--1.5)& $0.13\pm0.03$ & $1.00\pm0.04$ & $1.15\pm0.00$ & $-1.90\pm0.11$ & 0.34\\
\bottomrule
\end{tabular}
\caption{Fitted parameters for six of the nine evaluated language models (all nine are reported
per temperature in Table~\ref{tab:full}). Each entry is the mean $\pm$ sample standard deviation of
the per-temperature values in Table~\ref{tab:full} within each regime: Low $T\in\{0.0,0.1,0.2,0.3\}$,
Mid $T\in\{0.4,0.7\}$, High $T\in\{1.0,1.2,1.5\}$; $T=0$ denotes greedy decoding. $r$ is averaged
over per-temperature values, not recomputed from the averaged $\lambda,\epsilon$.
$\bar c\approx 1-r^2/2$ is the mean pairwise cosine similarity implied by $r$ for
$\ell_2$-normalized representations (Sec.~\ref{sec:limitations}).
$^{\dagger}$Per-temperature fits give $\lambda\in[-0.010,0.008]$, i.e.\ a flat, plateau-dominated
regime; negative values do not satisfy Definition~\ref{def:contraction} and are reported as $\lambda\approx0$, $r\approx\epsilon$.}
\label{tab:main}
\end{table}

\subsection{Image-Based Experiments}

We next test whether the same contraction--stochasticity parameterization describes iterative image generation, without claiming identical mechanisms across modalities.

\paragraph{Setup.}
We use 100 randomly sampled internet images resized to $512\times512$ as initial conditions. At each iteration, the image is shifted by an extension length $\mathrm{Ext}$ and the newly exposed region is generated using a pretrained inpainting model. We perform 100 iterations for $\mathrm{Ext}\in\{16,32,64,128\}$.

We evaluate Stable Diffusion v1.5 (SD v1.5)~\cite{rombach2022highresolution}, Stable Diffusion v2 (SD2)~\cite{stabilityai2022stablediffusion2}, and the RunwayML inpainting model~\cite{runwayml_inpainting}. As image representations we use the input activations of the final convolutional layer (\texttt{conv\_out}) of the denoising U-Net, the encoder--decoder network that predicts the noise in latent diffusion models. These features are recorded after each outpainting step, $\ell_2$-normalized, and analyzed using the same pairwise-distance recurrence as in the language experiments.

\begin{figure}[t]
\centering
\includegraphics[width=0.85\linewidth]{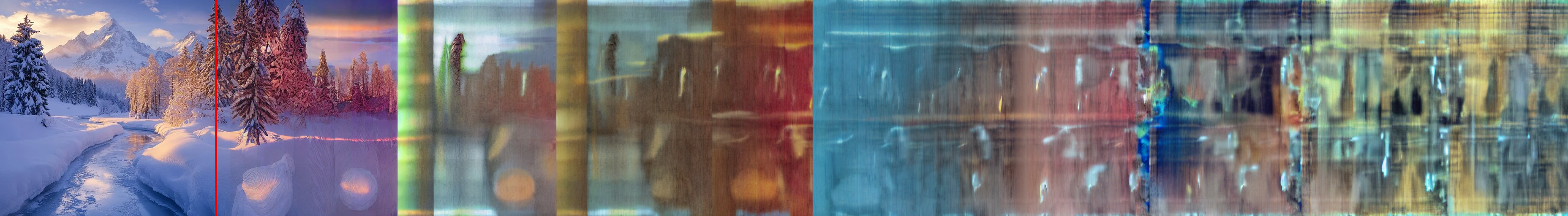}
\includegraphics[width=0.85\linewidth]{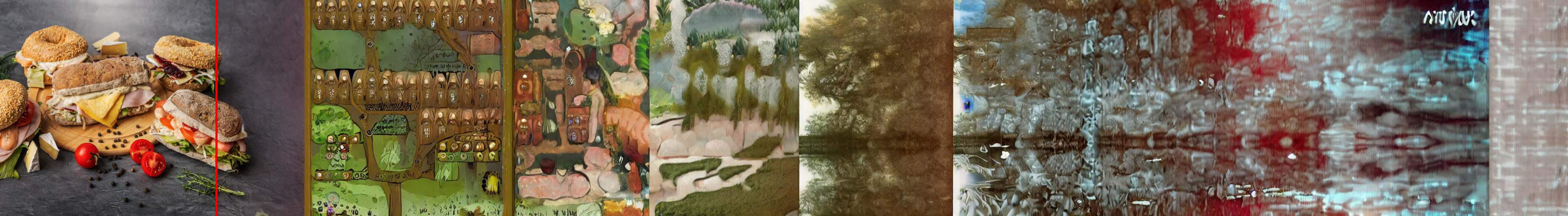}
\caption{
Example outpainting trajectories for Stable Diffusion v2 model with $\mathrm{Ext}=32$. The red line marks the boundary of the original image. Although the initial conditions differ substantially, the generated trajectories become progressively closer in representation space over the iterative generation process. The figure is illustrative of the trajectory geometry and does not directly measure perceptual output similarity.
}
\label{fig:image_examples}
\end{figure}

\paragraph{Trajectory dynamics.}
Representative trajectories are shown in Fig.~\ref{fig:image_examples}. Across substantially different initial images, pairwise representation distances decrease and subsequently fluctuate around a nonzero scale. As in language generation, this observation concerns representation-space geometry and does not establish convergence of image content or perceptual quality.

\paragraph{Estimated parameters.}
For each model and extension length, we estimate $\lambda$, $\epsilon$, $r$, and the noise-reduced Lyapunov-like diagnostic. We generate 100 trajectories per configuration, analyze 100 randomly selected trajectory pairs, and repeat the procedure five times. Uncertainties are computed across repetitions.

\begin{table}[t]
\footnotesize
\centering
\setlength{\tabcolsep}{4pt}
\begin{tabular}{l c c c c c c}
\toprule
Model & Ext & $\lambda$ & $\epsilon$ & $r$ & $\Lambda_{\mathrm{NR}}$ & $\bar c\approx1-r^2/2$ \\
\midrule
SD v1.5 & 16  & 0.3242 $\pm$ 0.01 & 0.8923 $\pm$ 0.06 & 1.3201 $\pm$ 0.36 & $-1.1267 \pm 0.09$ & 0.13\\
SD v1.5 & 32  & 0.6258 $\pm$ 0.08 & 0.4970 $\pm$ 0.02 & 1.3285 $\pm$ 0.39 & $-0.4688 \pm 0.16$ & 0.12\\
SD v1.5 & 64  & 0.6251 $\pm$ 0.08 & 0.4980 $\pm$ 0.03 & 1.3301 $\pm$ 0.39 & $-0.4700 \pm 0.15$ & 0.12\\
SD v1.5 & 128 & 0.6954 $\pm$ 0.11 & 0.4064 $\pm$ 0.03 & 1.3321 $\pm$ 0.37 & $-0.3632 \pm 0.22$ & 0.11\\
\midrule
SD2      & 16  & 0.6547 $\pm$ 0.15 & 0.4389 $\pm$ 0.08 & 1.2630 $\pm$ 0.22 & $-0.4237 \pm 0.18$ & 0.20\\
SD2      & 32  & 0.3397 $\pm$ 0.17 & 0.8561 $\pm$ 0.6 & 1.2927 $\pm$ 0.25 & $-1.0804 \pm 0.25$ & 0.16\\
SD2      & 64  & 0.3012 $\pm$ 0.13 & 0.9151 $\pm$ 0.4 & 1.3121 $\pm$ 0.31 & $-1.2010 \pm 0.29$ & 0.14\\
SD2      & 128 & 0.4527 $\pm$ 0.16 & 0.7142 $\pm$ 0.6 & 1.3043 $\pm$ 0.27 & $-0.7930 \pm 0.22$ & 0.15\\
\midrule
Runway   & 16  & 0.5702 $\pm$ 0.22 & 0.5665 $\pm$ 0.09 & 1.3150 $\pm$ 0.15 & $-0.5619 \pm 0.17$ & 0.14\\
Runway   & 32  & 0.5720 $\pm$ 0.28 & 0.5671 $\pm$ 0.11 & 1.3205 $\pm$ 0.16 & $-0.5586 \pm 0.16$ & 0.13\\
Runway   & 64  & 0.7953 $\pm$ 0.11 & 0.2718 $\pm$ 0.03 & 1.3264 $\pm$ 0.15 & $-0.2291 \pm 0.12$ & 0.12\\
Runway   & 128 & 0.8239 $\pm$ 0.18 & 0.2342 $\pm$ 0.02 & 1.3317 $\pm$ 0.18 & $-0.1937 \pm 0.14$ & 0.11\\
\bottomrule
\end{tabular}
\caption{
Estimated dynamical quantities across image-generation models and extension lengths. $\lambda$ is the fitted contraction coefficient, $\epsilon$ the effective stochastic scale, $r=\epsilon/(1-\lambda)$ the asymptotic separation scale, and $\Lambda_{\mathrm{NR}}$ the noise-reduced Lyapunov-like diagnostic. $\bar c\approx1-r^2/2$ is the mean pairwise cosine similarity implied by $r$ for $\ell_2$-normalized features; orthogonal unit vectors give $r=\sqrt2\approx1.41$ (Sec.~\ref{sec:limitations}). Uncertainties are computed across five repetitions.
}
\label{tab:image_dynamics}
\end{table}

\begin{figure}[t]
\centering
\small
\includegraphics[width=0.8\linewidth]{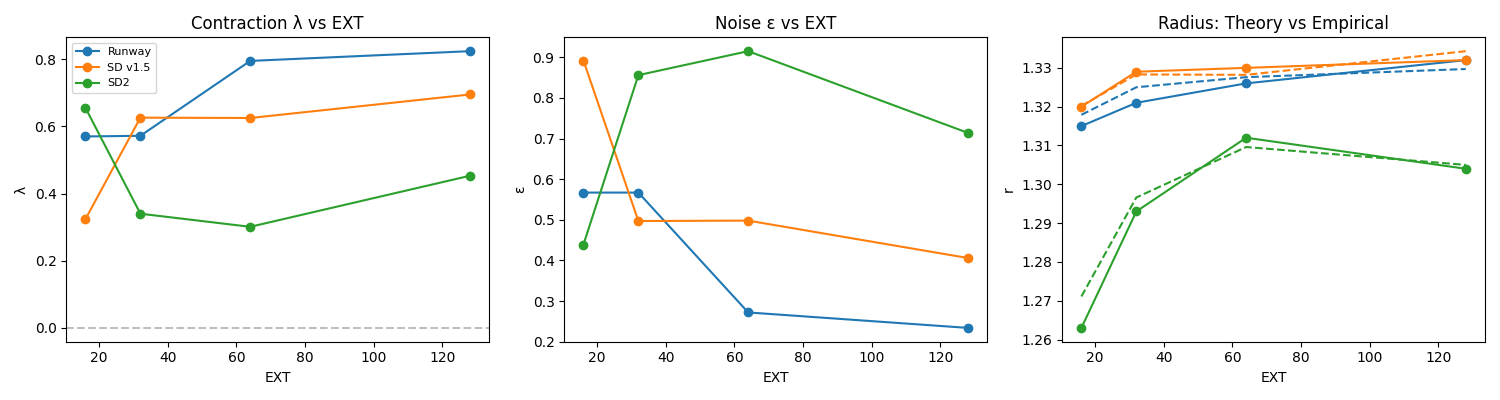}
\caption{
Effective dynamical parameters for diffusion-based image outpainting. Left: fitted contraction coefficient $\lambda$. Middle: effective stochastic scale $\epsilon$. Right: predicted asymptotic scale $r=\epsilon/(1-\lambda)$ compared with the observed long-horizon separation. Agreement in the right panel is largely implied by the least-squares fit (Sec.~\ref{sec:estimation}).
}
\label{fig:imgperturbation}
\end{figure}

Across image-generation settings, $\lambda$ and $\epsilon$ vary substantially, whereas $r$ remains comparatively stable, with values close to $1.3$ for most configurations. This corresponds to a mean cosine similarity of roughly $0.11$--$0.20$ between feature vectors of different trajectories, i.e.\ near-orthogonality (Sec.~\ref{sec:limitations}). Different combinations of contraction and stochasticity thus produce similar separation scales (Fig.~\ref{fig:imgperturbation}), although this stability may largely reflect feature-space geometry.

\paragraph{Dependence on extension length.}
The dependence on $\mathrm{Ext}$ is model-specific. For SD v1.5, larger extensions are associated with larger $\lambda$ and smaller $\epsilon$. SD2 exhibits its strongest contraction at intermediate extension lengths, while Runway shows increasing $\lambda$ and decreasing $\epsilon$ with extension length. Despite these differences, the resulting $r$ values remain similar across the three models. The outpainting experiments thus reproduce the qualitative pattern observed in language generation, supporting $(\lambda,\epsilon)$ as empirical coordinates for comparing long-horizon trajectory dynamics without implying a shared mechanism across modalities.

\section{Limitations}
\label{sec:limitations}

Several limitations constrain the interpretation of these results. First, the proposed stochastic dynamics are an effective description of observed trajectory statistics rather than a mechanistic model of the neural networks. The fitted $\lambda$ and $\epsilon$ are therefore not worst-case contraction guarantees or model-intrinsic constants, and the linear recurrence approximates potentially nonlinear, nonstationary, and state-dependent dynamics; its parameters may depend on the horizon, initialization, sampling procedure, and fitting interval. Because the fit is dominated by the plateau, $r$ largely reproduces the mean plateau distance (Sec.~\ref{sec:estimation}) and should be read as the scale predicted by the fitted model rather than an invariant radius of the underlying process.

Second, the results depend on the representation function and distance metric. We primarily use penultimate-layer representations for language models and pre-\texttt{conv\_out} U-Net features for outpainting, with Euclidean distance and cosine distance as a qualitative comparison. Other layers, normalizations, or metrics may yield different dynamical parameters.
In particular, since all representations are $\ell_2$-normalized, $\|u-v\|^2=2(1-\cos(u,v))$, so the separation scale corresponds to a mean pairwise cosine similarity $\bar c\approx1-r^2/2$ (up to a Jensen gap). Fig.~\ref{fig:perturbation} is consistent with this: the cosine-distance plateau ($\approx0.3$) and the Euclidean plateau ($\approx0.75$) satisfy $\sqrt{2\cdot0.3}\approx0.77$. For GPT-2, $r\leq0.26$ implies $\bar c\geq0.97$, in line with the extreme anisotropy of its upper layers reported by Ethayarajh~\cite{ethayarajh2019contextual}. Our results reproduce that finding, and the small GPT-2 separation scales largely reflect static representation geometry. For outpainting, $r\approx1.3$ implies $\bar c\approx0.11$--$0.20$, i.e.\ nearly orthogonal U-Net features (orthogonal unit vectors give $\sqrt2\approx1.41$). The stability of $r$ across image models and extension lengths may therefore be explained largely by high-dimensional feature geometry rather than by a specific balance between contraction and noise. We therefore interpret $r$ as a representation-dependent empirical scale. A static reference distance (e.g.\ between representations of unrelated, non-generated inputs) is needed before attributing differences in $r$ to dynamics.

Third, representation-space distance is only an indirect proxy for semantic similarity, contextual information, or output-level diversity. We do not report output-level measures such as repetition rate, distinct-$n$, or perceptual similarity, so the relation between $(\lambda,\epsilon,r)$ and output-level degeneration remains untested. Reduced hidden-state separation does not imply that generated outputs become equivalent or that meaningful information is lost. Establishing such a relationship requires token-, sequence-, or perceptual-level evaluation.

Finally, the evaluation covers a limited set of models and generation conditions, and contraction-like representation dynamics do not establish a causal mechanism for repetition, contextual degradation, or other forms of generation degeneration. Demonstrating causality would require interventions that modify the relevant dynamical properties and measure resulting changes in generation behavior. Our results instead provide an empirical framework for detecting and comparing contraction-like long-horizon dynamics.

\section{Conclusion}

We introduced an effective stochastic dynamical framework for characterizing long-horizon generative trajectories through pairwise representation-space distances. The framework summarizes trajectory evolution using an effective contraction coefficient $\lambda$ and stochastic scale $\epsilon$, which determine a relaxation timescale $\tau$ and long-horizon separation scale $r$. Across language and image-generation experiments, we observe a recurring pattern of decreasing representation-space separation followed by fluctuations around a nonzero long-horizon scale. The individual values of $\lambda$ and $\epsilon$ vary with decoding temperature, model architecture, and outpainting procedure, while their combined effect can yield comparatively stable values of $r$, which in turn partly reflect static representation geometry.

This provides a common empirical language for comparing long-horizon dynamics across generative systems without requiring strict attractors or unique limiting states.

Future work should extend the framework to state-dependent and time-varying contraction, alternative representation spaces, static reference distances that separate dynamics from representation geometry, output-level measures of diversity and semantic preservation, and controlled interventions that modify $(\lambda,\epsilon)$ and measure output-level effects, building on recent geometric interventions against mode collapse~\cite{du2026escaping}.

\section*{Acknowledgement}

This research was supported by the National Research, Development and Innovation Office (NKFIH), Hungary, under project no. 2026-4.1.1-MISSZIÓK-2026-00027. 

\bibliographystyle{abbrv}
\bibliography{llmcontraction}

@inproceedings{welleck2019neural,
  title={Neural text generation with unlikelihood training},
  author={Welleck, Sean and Kulikov, Ilia and Roller, Stephen and Dinan, Emily and Cho, Kyunghyun and Weston, Jason},
  booktitle={Proceedings of the International Conference on Learning Representations},
  year={2019}
}

@misc{runwayml_inpainting,
  author       = {{RunwayML}},
  title        = {Stable Diffusion Inpainting},
  year         = {2022},
  howpublished = {\url{https://huggingface.co/runwayml/stable-diffusion-inpainting}},
  note         = {Hugging Face model card}
}

@inproceedings{rombach2022highresolution,
  title={High-Resolution Image Synthesis with Latent Diffusion Models},
  author={Rombach, Robin and Blattmann, Andreas and Lorenz, Dominik and Esser, Patrick and Ommer, Bj{\"o}rn},
  booktitle={Proceedings of the IEEE/CVF Conference on Computer Vision and Pattern Recognition (CVPR)},
  year={2022}
}

@misc{stabilityai2022stablediffusion2,
  title={Stable Diffusion 2},
  author={{Stability AI}},
  year={2022},
  howpublished={\url{https://huggingface.co/stabilityai/stable-diffusion-2-inpainting}},
  note={Accessed: 2026-05-07}
}

@article{pham2009contraction,
  title={A contraction theory approach to stochastic incremental stability},
  author={Pham, Quang-Cuong and Tabareau, Nicolas and Slotine, Jean-Jacques},
  journal={IEEE Transactions on Automatic Control},
  volume={54},
  number={4},
  pages={816--820},
  year={2009},
  publisher={IEEE}
}

@article{diaconis1999iterated,
  title={Iterated random functions},
  author={Diaconis, Persi and Freedman, David},
  journal={SIAM review},
  volume={41},
  number={1},
  pages={45--76},
  year={1999},
  publisher={SIAM}
}

@inproceedings{du2026escaping,
  author={Xin Du and Kumiko Tanaka-Ishii},
  title={Escaping Mode Collapse in {LLM} Generation via Geometric Regulation},
  booktitle={International Conference on Machine Learning (ICML)},
  year={2026}, note={arXiv:2605.00435}}

@inproceedings{xu2022loop,
  author={Jin Xu and Xiaojiang Liu and Jianhao Yan and Deng Cai and Huayang Li and Jian Li},
  title={Learning to Break the Loop: Analyzing and Mitigating Repetitions for Neural Text Generation},
  booktitle={Advances in Neural Information Processing Systems}, year={2022}}

@inproceedings{xiao2024sinks,
  author={Guangxuan Xiao and Yuandong Tian and Beidi Chen and Song Han and Mike Lewis},
  title={Efficient Streaming Language Models with Attention Sinks},
  booktitle={International Conference on Learning Representations}, year={2024}}

@inproceedings{ethayarajh2019contextual,
  title={How contextual are contextualized word representations? Comparing the geometry of BERT, ELMo, and GPT-2 embeddings},
  author={Ethayarajh, Kawin},
  booktitle={Proceedings of the 2019 conference on empirical methods in natural language processing and the 9th international joint conference on natural language processing (EMNLP-IJCNLP)},
  pages={55--65},
  year={2019}
}

@article{kong2024representation,
  title={Aligning Large Language Models with Representation Editing: A Control Perspective},
  author={Kong, Y. and others},
  year={2024}
}

@article{zhang2026closing,
  title={Closing the Loop: A Control-Theoretic Framework for Provably Stable Time Series Forecasting with LLMs},
  author={Zhang, Xingyu and Du, Hanyun and Song, Zeen and Zhang, Jianqi and Zheng, Changwen and Qiang, Wenwen},
  journal={arXiv preprint arXiv:2602.12756},
  year={2026}
}

@article{haber2018stable,
  title={Stable architectures for deep neural networks},
  author={Haber, Eldad and Ruthotto, Lars},
  journal={Inverse problems},
  volume={34},
  number={1},
  pages={014004},
  year={2018},
  publisher={IOP Publishing}
}

@article{gomez2017reversible,
  title={The reversible residual network: Backpropagation without storing activations},
  author={Gomez, Aidan N and Ren, Mengye and Urtasun, Raquel and Grosse, Roger B},
  journal={Advances in neural information processing systems},
  volume={30},
  year={2017}
}

@article{bai2019deep,
  title={Deep equilibrium models},
  author={Bai, Shaojie and Kolter, J Zico and Koltun, Vladlen},
  journal={Advances in neural information processing systems},
  volume={32},
  year={2019}
}

@article{chang2017multi,
  title={Multi-level residual networks from dynamical systems view},
  author={Chang, Bo and Meng, Lili and Haber, Eldad and Tung, Frederick and Begert, David},
  journal={arXiv preprint arXiv:1710.10348},
  year={2017}
}

@article{naparstek2026token,
  title={Token Maturation: Autoregressive Language Generation via Continuous Token Dynamics},
  author={Naparstek, Oshri},
  journal={arXiv preprint arXiv:2601.04854},
  year={2026}
}

@inproceedings{wang2023shall,
  title={Shall we pretrain autoregressive language models with retrieval? a comprehensive study},
  author={Wang, Boxin and Ping, Wei and Xu, Peng and McAfee, Lawrence and Liu, Zihan and Shoeybi, Mohammad and Dong, Yi and Kuchaiev, Oleksii and Li, Bo and Xiao, Chaowei and others},
  booktitle={Proceedings of the 2023 conference on empirical methods in natural language processing},
  pages={7763--7786},
  year={2023}
}

@inproceedings{khandelwal2018sharp,
  title={Sharp nearby, fuzzy far away: How neural language models use context},
  author={Khandelwal, Urvashi and He, He and Qi, Peng and Jurafsky, Dan},
  booktitle={Proceedings of the 56th annual meeting of the Association for Computational Linguistics (Volume 1: Long Papers)},
  pages={284--294},
  year={2018}
}

@article{holtzman2019curious,
  title={The curious case of neural text degeneration},
  author={Holtzman, Ari and Buys, Jan and Du, Li and Forbes, Maxwell and Choi, Yejin},
  journal={arXiv preprint arXiv:1904.09751},
  year={2019}
}

@article{press2021train,
  title={Train short, test long: Attention with linear biases enables input length extrapolation},
  author={Press, Ofir and Smith, Noah A and Lewis, Mike},
  journal={arXiv preprint arXiv:2108.12409},
  year={2021}
}

@inproceedings{fan2018hierarchical,
  title={Hierarchical neural story generation},
  author={Fan, Angela and Lewis, Mike and Dauphin, Yann},
  booktitle={Proceedings of the 56th Annual Meeting of the Association for Computational Linguistics (Volume 1: Long Papers)},
  pages={889--898},
  year={2018}
}

@article{chen2018neural,
  title={Neural ordinary differential equations},
  author={Chen, Ricky TQ and Rubanova, Yulia and Bettencourt, Jesse and Duvenaud, David K},
  journal={Advances in neural information processing systems},
  volume={31},
  year={2018}
}

@article{liu2024lost,
  title={Lost in the middle: How language models use long contexts},
  author={Liu, Nelson F and Lin, Kevin and Hewitt, John and Paranjape, Ashwin and Bevilacqua, Michele and Petroni, Fabio and Liang, Percy},
  journal={Transactions of the association for computational linguistics},
  volume={12},
  pages={157--173},
  year={2024}
}

@article{wiher2022decoding,
  title={On decoding strategies for neural text generators},
  author={Wiher, Gian and Meister, Clara and Cotterell, Ryan},
  journal={Transactions of the Association for Computational Linguistics},
  volume={10},
  pages={997--1012},
  year={2022},
  publisher={MIT Press One Broadway, 12th Floor, Cambridge, Massachusetts 02142, USA~…}
}

\appendix \section*{Appendix} \addcontentsline{toc}{section}{Appendix}

\section{Detailed proofs}

The arguments below are standard and follow the stochastic contraction analysis of Pham, Tabareau and Slotine~\cite{pham2009contraction} and the iterated-random-function analysis of Diaconis and Freedman~\cite[Sec.~5]{diaconis1999iterated}; we reproduce them for completeness.

\subsection{Proof of Theorem~\ref{thm:distance}}

\begin{theorem*}[Theorem~\ref{thm:distance}, restated]
Under Definition~\ref{def:contraction},
\begin{equation}
\mathbb{E}\,\|z_t - z_t'\| \le \lambda^t\, \mathbb{E}\,\|z_0 - z_0'\| + \frac{\epsilon}{1-\lambda}(1 - \lambda^t).
\end{equation}
If the initial states are deterministic, $\mathbb{E}\,\|z_0 - z_0'\| = \|z_0 - z_0'\|$.
\end{theorem*}

\begin{proof}
Let
\[
d_t := \mathbb{E}\,\|z_t - z_t'\|.
\]
By Definition~\ref{def:contraction}, the sequence $(d_t)$ satisfies the recursive inequality
\begin{equation}
d_{t+1} \le \lambda d_t + \epsilon.
\label{eq:recursion}
\end{equation}

We prove by induction that
\begin{equation}
d_t \le \lambda^t d_0 + \epsilon \sum_{k=0}^{t-1} \lambda^k.
\label{eq:unrolled}
\end{equation}

\paragraph{Base case ($t=0$).}
The claim holds trivially:
\[
d_0 = \lambda^0 d_0 + 0.
\]

\paragraph{Inductive step.}
Assume \eqref{eq:unrolled} holds for some $t \ge 0$. Then using \eqref{eq:recursion},
\begin{align*}
d_{t+1}
&\le \lambda d_t + \epsilon \\
&\le \lambda \left( \lambda^t d_0 + \epsilon \sum_{k=0}^{t-1} \lambda^k \right) + \epsilon \\
&= \lambda^{t+1} d_0 + \epsilon \sum_{k=1}^{t} \lambda^k + \epsilon \\
&= \lambda^{t+1} d_0 + \epsilon \left( \sum_{k=1}^{t} \lambda^k + 1 \right) \\
&= \lambda^{t+1} d_0 + \epsilon \sum_{k=0}^{t} \lambda^k.
\end{align*}
This matches the desired form for $t+1$, completing the induction.

\paragraph{Conclusion.}
Using the geometric series identity
\[
\sum_{k=0}^{t-1} \lambda^k = \frac{1 - \lambda^t}{1 - \lambda},
\]
we obtain
\[
d_t \le \lambda^t d_0 + \frac{\epsilon}{1-\lambda}(1 - \lambda^t),
\]
which proves the result.
\end{proof}

\paragraph{Short-horizon regime ($t \ll \tau$).}
Let $d_t := \mathbb{E}\,\|z_t - z_t'\|$. From Theorem~\ref{thm:distance},
\[
d_t \le \lambda^t d_0 + \frac{\epsilon}{1-\lambda}(1 - \lambda^t).
\]
For $\lambda$ close to $1$, we can write $\lambda^t \approx e^{-t(1-\lambda)} = e^{-t/\tau}$, where $\tau = \frac{1}{1-\lambda}$. In the regime $t \ll \tau$, a first-order expansion yields
\[
\lambda^t \approx 1 - \frac{t}{\tau}.
\]
Substituting into the bound gives
\[
d_t \lesssim d_0 \left(1 - \frac{t}{\tau}\right) + \epsilon t.
\]
Thus, the contribution from the initial condition remains dominant, while the additive term accumulates approximately linearly in time. In particular, when distances are small, the additive term can dominate, which is compatible with transient increases in $d_t$ and apparent expansion.

\paragraph{Long-horizon regime ($t \gg \tau$).}
In the regime $t \gg \tau$, we have $\lambda^t \to 0$, and the bound simplifies to
\[
d_t \le \frac{\epsilon}{1-\lambda}(1 - \lambda^t) \to \frac{\epsilon}{1-\lambda}.
\]
At the same time, the dependence on the initial condition vanishes:
\[
\lambda^t d_0 \to 0.
\]
Therefore, the bound becomes independent of initialization and approaches the characteristic scale
\[
r = \frac{\epsilon}{1 - \lambda}.
\]
This yields an upper bound on expected long-horizon separation determined by the balance between contraction and the additive term. We note that this establishes an asymptotic upper bound, not convergence of $d_t$ to $r$ or of the trajectories to a common state.

\subsection{Proof of Eq.~\eqref{eq:asymptotic_bound}}

Theorem~\ref{thm:distance} implies an asymptotic upper bound on expected separation (cf.~\cite[Eq.~2.13]{pham2009contraction},~\cite[Thm.~4.1]{zhang2026closing}).

\begin{corollary}[Asymptotic separation bound]
\label{cor:asymptotic}
Under Definition~\ref{def:contraction},
\[
\limsup_{t \to \infty} \mathbb{E}\,\|z_t - z_t'\| \le \frac{\epsilon}{1 - \lambda}.
\]
\end{corollary}

\begin{proof}
Let
\[
d_t := \mathbb{E}\,\|z_t - z_t'\|.
\]
By Theorem~\ref{thm:distance}, we have the bound
\begin{equation}
d_t \le \lambda^t d_0 + \frac{\epsilon}{1-\lambda}(1 - \lambda^t).
\label{eq:distance_bound_recall}
\end{equation}

We now analyze the asymptotic behavior of the right-hand side. Since $\lambda \in [0,1)$, we have
\[
\lambda^t \to 0 \quad \text{as } t \to \infty.
\]
Therefore,
\[
\lambda^t d_0 \to 0,
\qquad
1 - \lambda^t \to 1.
\]

Taking the limit superior on both sides of \eqref{eq:distance_bound_recall} and using standard properties of $\limsup$, we obtain
\begin{align*}
\limsup_{t \to \infty} d_t
&\le \limsup_{t \to \infty} \left( \lambda^t d_0 + \frac{\epsilon}{1-\lambda}(1 - \lambda^t) \right) \\
&= \lim_{t \to \infty} \lambda^t d_0
\;+\;
\lim_{t \to \infty} \frac{\epsilon}{1-\lambda}(1 - \lambda^t) \\
&= 0 + \frac{\epsilon}{1-\lambda}.
\end{align*}

Thus,
\[
\limsup_{t \to \infty} \mathbb{E}\,\|z_t - z_t'\|
\le
\frac{\epsilon}{1-\lambda},
\]
which proves the result.
\end{proof}

\section{Detailed description of experiments}

\subsection{Environment}
All experiments were conducted on a single-GPU system equipped with 12 GB of GPU memory, using standard software frameworks. All code, together with the generated and saved trajectories, will be made publicly available.

\subsection{Text-based experiments}

\paragraph{Corpus construction.}
We construct a dataset of $1500$ text prompts drawn from three sources: Wikipedia ($40\%$), news articles ($40\%$), and synthetic text ($20\%$), ensuring diversity in both topic and style. All samples are constrained to a fixed length range: texts shorter than $500$ characters are discarded, while longer texts are truncated to a maximum of $4000$ characters, providing sufficiently informative yet controlled inputs.

Wikipedia data are obtained via the public REST API by sampling random page titles and retrieving the corresponding article content. The main text is extracted by parsing paragraph elements and concatenating them into a single sequence. Articles that fail retrieval or do not meet the minimum length requirement are discarded, with multiple retry attempts allowed; if the target number of samples is not reached, fallback generic text is used to ensure dataset completeness. News articles are collected from RSS feeds of major outlets (specifically BBC and CNN). For each entry, the linked webpage is fetched and parsed, and the article text is extracted from paragraph tags. Samples that fail parsing or length constraints are skipped, with fallback text again used if necessary to reach the required count.

Synthetic samples are generated using templated text based on a small set of predefined topics (including artificial intelligence, climate change, economics, space exploration, history of civilizations, and philosophy of mind). For each sample, a topic is selected uniformly at random and expanded into a repeated descriptive template, producing texts of comparable length to real data and providing controlled, low-entropy inputs.

All texts are normalized by removing line breaks and collapsing whitespace. The combined dataset is deduplicated and randomly shuffled to eliminate ordering effects. After deduplication, the first 1000 samples are used. The final corpus is stored as a plain text file with samples separated by blank lines, and metadata recording the number of samples per source is saved separately. This procedure yields a fixed-size dataset with balanced composition, controlled length distribution, and consistent formatting, providing stable initial conditions for subsequent trajectory analysis.

\paragraph{Trajectory generation.}
For each prompt in the corpus, we generate autoregressive trajectories using nine pretrained causal language models: GPT-2, Gemma-2B, Gemma-2-2B-it, Llama-3.2-1B, Llama-3.2-1B-Instruct, Qwen2.5-3B, Qwen2.5-3B-Instruct, TinyLlama-1.1B-Chat-v1.0, and Phi-2 (Gemma-2B in Table~\ref{tab:main} refers to the base model gemma-2b). For each model, we sweep temperature values $T \in \{0.0, 0.1, 0.2, 0.3, 0.4, 0.7, 1.0, 1.2, 1.5\}$ while keeping other decoding parameters fixed.

Generation proceeds for $1000$ autoregressive steps, producing one token per step. For $T=0$ we use greedy decoding; for $T>0$, sampling is performed using temperature scaling followed by top-$k$ truncation with $k=40$, after which tokens are drawn from the resulting distribution. Numerical stabilization is applied prior to softmax to ensure well-behaved probabilities. No early stopping is used, and end-of-sequence tokens are not enforced, ensuring all trajectories have equal length.

At each generation step, we extract the hidden representation corresponding to the final token from the penultimate model layer. These representations define the trajectory $(z_t)$ in $\mathbb{R}^d$. To reduce storage and computational overhead, representations are recorded every $10$ steps, yielding a subsampled trajectory while preserving long-horizon dynamics. All representations are stored in half-precision format.

Each trajectory is initialized from a truncated version of the input prompt: specifically, the first $400$ characters are retained and cut to the last complete sentence to ensure coherent starting conditions. A fixed instruction prefix is prepended to all inputs to standardize generation behavior across models.

To analyze sensitivity to initial conditions, we generate multiple perturbed versions of each prompt. For every base input, $10$ additional variants are constructed using a simple token-level perturbation: two randomly selected words in the text are swapped. This produces minimally altered inputs that preserve overall semantics while introducing small perturbations in representation space. All perturbed and unperturbed inputs are processed independently, yielding a set of trajectories for each original prompt.

Generation is performed using standard transformer caching (key-value cache) for efficiency. Unless otherwise specified, the full context is retained throughout generation (i.e., no truncation is applied), ensuring that each step conditions on the entire generated history. Optional experiments with sliding context windows were supported but not used in the main results.

For each trajectory, we store (i) the sequence of extracted representations, (ii) the final generated text, and (iii) metadata including runtime and configuration parameters. Experiments are organized by model, temperature, perturbation type, and random seed. This procedure yields a large collection of trajectories with controlled initialization and stochasticity, enabling systematic analysis of contraction dynamics.

\paragraph{Distance computation.}
For each experiment configuration (defined by model and temperature), we aggregate trajectories generated from all prompts and their perturbations. Each trajectory consists of a time series of hidden representations $(z_t)$. Prior to any computation, all representations are $\ell_2$-normalized at each time step to remove scale effects, so that distances reflect angular differences in representation space; in particular, $\|u-v\|^2 = 2(1-\cos(u,v))$ for normalized $u,v$. Trajectories are temporally aligned by truncating them to a common minimum length across all samples. Invalid or non-finite values are discarded, and missing values are filled using linear interpolation to ensure a continuous curve. Two kinds of trajectory pairs are used:
(i) \emph{baseline pairs}, i.e.\ $1000$ randomly sampled pairs of unperturbed trajectories from different prompts, used to estimate $(\lambda,\epsilon)$; and
(ii) \emph{baseline--perturbed pairs}, used for the Lyapunov-like diagnostics. For the latter, at each time step $t$ the distance is the average Euclidean distance between the normalized base representation and all corresponding perturbed representations:
\[
d_t = \mathbb{E}_{\text{pert}} \left[ \| z_t - z_t^{(\text{pert})} \|_2 \right].
\]

\paragraph{Estimation of $(\lambda,\epsilon)$.}
Using the baseline pairs, we fit the relation
\[
d_{t+1} \approx \lambda d_t + \epsilon
\]
by linear regression on the mean distance curve. The slope $\lambda$ is the effective contraction coefficient, and the intercept $\epsilon$ is the additive term, from which $r = \epsilon/(1-\lambda)$ follows. Because the fit covers the full horizon, which is dominated by the plateau, $r$ largely reproduces the mean plateau distance (Sec.~\ref{sec:estimation}). In addition to this global fit, we compute local estimates using a sliding window of size $10$ to capture temporal variation in $(\lambda, \epsilon)$. Within each window, we also compute the Pearson correlation between $d_t$ and $d_{t+1}$ to check the linear relationship.

\paragraph{Lyapunov-like diagnostics.}
Using the baseline--perturbed pairs, the global exponent is computed as
\[
\Lambda_{\text{global}} = \frac{1}{T} \log \frac{d_T}{d_0},
\]
while an early-time exponent is computed over the first $5$ recorded steps to capture short-term behavior. A local exponent is additionally estimated as the average of $\log(d_{t+1}/d_t)$ over all time steps, providing a step-wise characterization of expansion or contraction. To reduce the influence of the additive term, we compute a noise-reduced exponent from pairs of perturbed trajectories. For each such pair $(z_t, z_t')$, we compute distances $d_t$ and apply a first-order correction
\[
\tilde{d}_{t+1} = d_{t+1} - \epsilon,
\]
where $\epsilon$ is the value estimated above. The corrected growth rates $\log(\tilde{d}_{t+1}/d_t)$ are averaged across all valid time steps and trajectory pairs to obtain $\Lambda_{\mathrm{NR}}$. This correction removes a first-order additive bias; because it uses the fitted $\epsilon$, $\Lambda_{\mathrm{NR}}$ is not independent of $\log\lambda$ (Sec.~\ref{sec:estimation}). Finally, all quantities are aggregated across prompts for each model and temperature setting, yielding a single set of parameters $(\lambda, \epsilon)$ and Lyapunov estimates per configuration.

\subsection{Image-based experiments}

\paragraph{Trajectory generation.}
We construct an iterative outpainting process that mirrors autoregressive generation in a continuous spatial domain. Each experiment begins from an initial image $x_0 \in \mathbb{R}^{512 \times 512 \times 3}$, drawn from a set of 100 diverse natural images, which are resized to a fixed resolution of $512 \times 512$. Given a pretrained diffusion-based inpainting model, we define a sequence $(x_t)_{t=0}^T$ with $T = 100$ by repeatedly extending the image along the horizontal axis. At each step, the current global canvas has width $512 + t \cdot \mathrm{Ext}$, where the extension parameter $\mathrm{Ext} \in \{16,32,64,128\}$ controls the number of newly generated pixels per iteration. The dynamics proceed by selecting a sliding window of fixed size $512 \times 512$ aligned with the current generation frontier. Formally, if the frontier position is $f_t = t \cdot \mathrm{Ext}$, we extract a crop $c_t = x_t[f_t : f_t + 512, \, 0 : 512]$. A binary mask $m_t \in \{0,1\}^{512 \times 512}$ is constructed such that the rightmost $\mathrm{Ext}$ columns are masked (set to 1) and the remaining pixels are unmasked (set to 0). The next state is obtained via the inpainting operator
\[
x_{t+1} = F_\theta(c_t, m_t; \eta_t),
\]
where $F_\theta$ denotes the diffusion inpainting model and $\eta_t$ captures the stochasticity of the denoising process. Generation is performed using classifier-free guidance with scale $7.5$ and $25$ inference steps, and a fixed prompt (``continuous natural scene, coherent extension, same style'') to encourage consistency across iterations. After generation, only the newly synthesized region of width $\mathrm{Ext}$ is appended back to the global canvas, so that information propagates only through previously generated content, analogous to autoregressive dependence. We perform this procedure across three models (Stable Diffusion v1.5 inpainting, Stable Diffusion 2 inpainting, and RunwayML inpainting), for all values of $\mathrm{Ext}$ and all input images.

To analyze the induced dynamics in representation space, we extract features at each time step by registering a forward hook on the final convolutional layer of the U-Net (\texttt{conv\_out}), the layer that maps the U-Net's last feature maps to the predicted noise in latent space. We record the input activations to this layer, denoted $z_t \in \mathbb{R}^d$, as a continuous representation of the system state. These features are stored at every iteration, producing trajectories $(z_t)_{t=0}^T$ that are directly comparable across models and settings.

\paragraph{Distance computation and parameter estimation.}
Given a set of trajectories $(z_t^{(i)})_{t=0}^T$, we first flatten the extracted feature tensors into vectors in $\mathbb{R}^d$ and apply $\ell_2$ normalization at every time step, i.e., $z_t \leftarrow z_t / (\|z_t\| + \varepsilon)$ with a small constant $\varepsilon = 10^{-8}$ for numerical stability. Trajectories containing invalid values (NaNs, infinities, or near-zero norm vectors) or with fewer than two valid time steps are discarded. From the remaining set of trajectories (at most 100 per configuration), we uniformly sample $N = 1000$ random pairs $(i,j)$ and compute their distance evolution over time as
\[
d_t^{(i,j)} = \|z_t^{(i)} - z_t^{(j)}\|_2.
\]
To fit the contractive model
\[
d_{t+1} \approx \lambda d_t + \epsilon,
\]
we aggregate all valid pairs and time steps and perform a global least-squares regression with $x = d_t$, $y = d_{t+1}$, yielding a single pair $(\lambda,\epsilon)$ per model and extension setting. From these, we compute the timescale $\tau = \frac{1}{1-\lambda}$ and the predicted separation scale $r_{\text{theory}} = \frac{\epsilon}{1-\lambda}$ (for $\lambda < 1$). We also compute the empirical long-horizon separation by averaging distances over the final portion of each trajectory,
\[
r_{\text{emp}} = \mathbb{E}_{i,j}\left[ \frac{1}{|T'|} \sum_{t \in T'} d_t^{(i,j)} \right],
\]
where $T'$ corresponds to the last $20\%$ of time steps. Since the regression is dominated by plateau samples, close agreement between $r_{\text{theory}}$ and $r_{\text{emp}}$ is largely implied by the fit (Sec.~\ref{sec:estimation}).

To characterize local expansion properties, we compute a (naïve) Lyapunov-like exponent from trajectory pairs whose initial inputs are an original image and a perturbed version of it:
\[
\Lambda = \mathbb{E}\left[ \log \frac{d_{t+1}}{d_t} \right],
\]
averaged over all valid pairs and time steps where distances are strictly positive and finite. Since this estimate is biased by the additive term, we also compute a noise-corrected exponent by subtracting the estimated $\epsilon$ from the numerator,
\[
\Lambda_{\text{NR}} = \mathbb{E}\left[ \log \frac{d_{t+1} - \epsilon}{d_t} \right],
\]
again restricting to valid regions where both numerator and denominator remain positive. All statistics are computed by aggregating over the sampled $N=1000$ trajectory pairs.

\section{Contraction in other layers}

To examine how the fitted parameters depend on depth, we estimate them using representations extracted at different network layers. These measurements do not correspond to the local action of a single layer. Instead, the representation at layer $k$ reflects the composition of all preceding transformations, i.e., the mapping induced by layers $1, \dots, k$ applied to the input sequence. Each estimate at depth $k$ therefore describes the distance dynamics of representations produced by the first $k$ layers.

\begin{figure}[!hpt]
\centering
 \small
\includegraphics[width=0.9\linewidth]{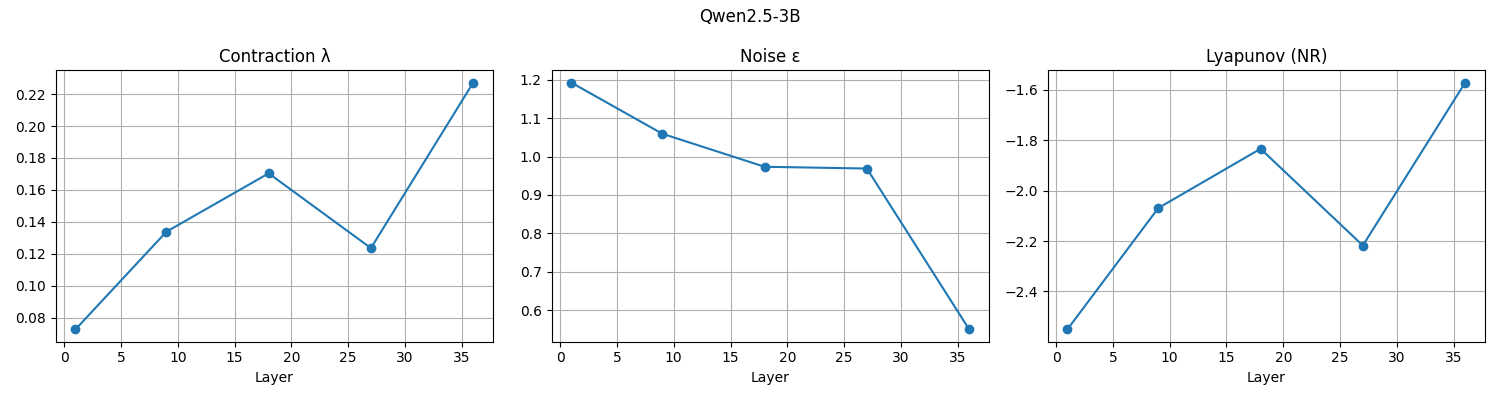}
\caption{
Layerwise estimates of contraction ($\lambda$), additive term ($\epsilon$), and noise-reduced Lyapunov-like exponent for Qwen2.5-3B at temperature $0.4$.
Representations from early layers show stronger fitted contraction (lower $\lambda$), larger $\epsilon$, and more negative $\Lambda_{\mathrm{NR}}$.
Deeper layers show weaker contraction and smaller $\epsilon$, i.e.\ a weakly contracting regime with slower fitted decay. All values of $\lambda$ remain well below $1$.
}
\label{fig:layerwiseperturbation}
\end{figure}

\paragraph{Layerwise structure.}
We observe a systematic depth-dependent structure in the fitted parameters (Fig.~\ref{fig:layerwiseperturbation}). Representations from early layers show stronger fitted contraction (smaller $\lambda$) combined with a larger additive term $\epsilon$, while representations from deeper layers show weaker contraction (larger $\lambda$) and a smaller $\epsilon$. Over all layers, $\lambda$ remains small (approximately $0.08$--$0.22$), so no layer is close to the non-contracting regime $\lambda \approx 1$. Early-layer representations also have more negative $\Lambda_{\mathrm{NR}}$ values. Whether this corresponds to a loss of information about initial conditions would require probing experiments, which we leave for future work.

\paragraph{Implications.}
Within the fitted model, the strongest contraction is observed when distances are measured at early layers, while deeper layers operate within a bounded region of representation space with weaker contraction. One possible interpretation, which we leave for future work, is that early layers contribute most of the reduction in trajectory separation while later layers apply more structure-preserving transformations. As in the main text, these are descriptive statistics of the chosen representations and do not establish a mechanism.

\section{Detailed results from our experiments}

We evaluated a large set of models and temperatures. For readability, the main text reports regime-averaged values for six models (Table~\ref{tab:main}); the full per-temperature results for all nine models are given in Table~\ref{tab:full}.

\begin{longtable}{lccccc}
\caption{Full parameter table grouped by model and ordered by temperature ($T=0$: greedy decoding). The regression is used as an empirical estimator of the recurrence coefficient. In noise-dominated or nearly flat regimes, an unconstrained least-squares fit can produce a small negative slope. Such values do not satisfy Definition~\ref{def:contraction} and are therefore not interpreted as contraction coefficients; they indicate that the corresponding empirical distance evolution is effectively flat, with $\lambda\approx0$ and $r\approx\epsilon$. Table~\ref{tab:main} averages these values within temperature regimes.}
\label{tab:full}\\
\toprule
Model & Temp & $\lambda$ & $\epsilon$ & $r=\frac{\epsilon}{1-\lambda}$ & $\Lambda_{\mathrm{NR}}$ \\
\midrule
\endfirsthead

\toprule
Model & Temp & $\lambda$ & $\epsilon$ & $r$ & $\Lambda_{\mathrm{NR}}$ \\
\midrule
\endhead

\midrule
\multicolumn{6}{c}{\textbf{gemma-2b}}\\
\midrule
gemma-2b & 0.0 & 0.4973 & 0.1514 & 0.3010 & -1.3320 \\
gemma-2b & 0.1 & 0.9014 & 0.0602 & 0.6101 & -0.2266 \\
gemma-2b & 0.2 & 0.6748 & 0.3589 & 1.1038 & -0.6257 \\
gemma-2b & 0.3 & 0.2042 & 0.5944 & 0.7471 & -1.4265 \\
gemma-2b & 0.4 & 0.1683 & 0.6000 & 0.7213 & -1.4411 \\
gemma-2b & 0.7 & 0.0727 & 0.6248 & 0.6736 & -1.4719 \\
gemma-2b & 1.0 & 0.0877 & 0.6103 & 0.6690 & -1.4536 \\
gemma-2b & 1.2 & 0.0450 & 0.6403 & 0.6704 & -1.4444 \\
gemma-2b & 1.5 & 0.1314 & 0.5828 & 0.6710 & -1.4806 \\

\midrule
\multicolumn{6}{c}{\textbf{gemma-2-2b-it}}\\
\midrule
gemma-2-2b-it & 0.0 & 0.5794 & 0.1828 & 0.4346 & -0.8056 \\
gemma-2-2b-it & 0.1 & 0.4640 & 0.2422 & 0.4520 & -1.0986 \\
gemma-2-2b-it & 0.2 & 0.5276 & 0.2371 & 0.5020 & -1.0068 \\
gemma-2-2b-it & 0.3 & 0.7144 & 0.1764 & 0.6175 & -0.5452 \\
gemma-2-2b-it & 0.4 & 0.4508 & 0.4923 & 0.8969 & -0.8584 \\
gemma-2-2b-it & 0.7 & 0.0749 & 0.8363 & 0.9041 & -1.8241 \\
gemma-2-2b-it & 1.0 & 0.0329 & 0.8510 & 0.8799 & -1.8753 \\
gemma-2-2b-it & 1.2 & 0.0529 & 0.8307 & 0.8771 & -1.8806 \\
gemma-2-2b-it & 1.5 & 0.0645 & 0.8258 & 0.8831 & -1.8873 \\

\midrule
\multicolumn{6}{c}{\textbf{gpt2}}\\
\midrule
gpt2 & 0.0 & 0.7884 & 0.0140 & 0.0660 & -0.3703 \\
gpt2 & 0.1 & 0.8227 & 0.0135 & 0.0762 & -0.3172 \\
gpt2 & 0.2 & 0.7126 & 0.0406 & 0.1413 & -0.5708 \\
gpt2 & 0.3 & 0.3522 & 0.1329 & 0.2053 & -1.2398 \\
gpt2 & 0.4 & 0.2019 & 0.1866 & 0.2337 & -1.2744 \\
gpt2 & 0.7 & -0.0288 & 0.2619 & 0.2546 & -1.0153 \\
gpt2 & 1.0 & -0.0070 & 0.2625 & 0.2606 & -0.9787 \\
gpt2 & 1.2 & 0.0082 & 0.2562 & 0.2583 & -1.0479 \\
gpt2 & 1.5 & -0.0096 & 0.2645 & 0.2619 & -1.0167 \\

\midrule
\multicolumn{6}{c}{\textbf{Llama-3.2-1B}}\\
\midrule
Llama-3.2-1B & 0.0 & 0.7362 & 0.1503 & 0.5697 & -0.3552 \\
Llama-3.2-1B & 0.1 & 0.7841 & 0.1305 & 0.6035 & -0.2711 \\
Llama-3.2-1B & 0.2 & 0.7031 & 0.2197 & 0.7396 & -0.3769 \\
Llama-3.2-1B & 0.3 & 0.4035 & 0.5091 & 0.8537 & -1.0038 \\
Llama-3.2-1B & 0.4 & 0.3588 & 0.5680 & 0.8859 & -1.1551 \\
Llama-3.2-1B & 0.7 & 0.3078 & 0.6214 & 0.8979 & -1.4306 \\
Llama-3.2-1B & 1.0 & 0.2683 & 0.6708 & 0.9167 & -1.6044 \\
Llama-3.2-1B & 1.2 & 0.2468 & 0.6843 & 0.9085 & -1.7052 \\
Llama-3.2-1B & 1.5 & 0.2998 & 0.6374 & 0.9103 & -1.4716 \\

\midrule
\multicolumn{6}{c}{\textbf{Llama-3.2-1B-Instruct}}\\
\midrule
Llama-3.2-1B-Instruct & 0.0 & 0.3634 & 0.4635 & 0.7284 & -1.0291 \\
Llama-3.2-1B-Instruct & 0.1 & 0.4161 & 0.4481 & 0.7673 & -0.9456 \\
Llama-3.2-1B-Instruct & 0.2 & 0.2888 & 0.6081 & 0.8549 & -1.3718 \\
Llama-3.2-1B-Instruct & 0.3 & 0.2014 & 0.7192 & 0.9004 & -1.7921 \\
Llama-3.2-1B-Instruct & 0.4 & 0.1076 & 0.8148 & 0.9132 & -2.1535 \\
Llama-3.2-1B-Instruct & 0.7 & 0.1336 & 0.7873 & 0.9085 & -2.1211 \\
Llama-3.2-1B-Instruct & 1.0 & 0.0994 & 0.8052 & 0.8943 & -2.2176 \\
Llama-3.2-1B-Instruct & 1.2 & 0.1038 & 0.8079 & 0.9020 & -2.2189 \\
Llama-3.2-1B-Instruct & 1.5 & 0.1420 & 0.7692 & 0.8963 & -2.0877 \\

\midrule
\multicolumn{6}{c}{\textbf{Qwen2.5-3B}}\\
\midrule
Qwen2.5-3B & 0.0 & 0.3944 & 0.3074 & 0.5079 & -1.0390 \\
Qwen2.5-3B & 0.1 & 0.4935 & 0.2696 & 0.5323 & -0.8050 \\
Qwen2.5-3B & 0.2 & 0.1637 & 0.5465 & 0.6534 & -1.7281 \\
Qwen2.5-3B & 0.3 & 0.2942 & 0.4894 & 0.6932 & -1.4853 \\
Qwen2.5-3B & 0.4 & 0.2800 & 0.5131 & 0.7127 & -1.4895 \\
Qwen2.5-3B & 0.7 & 0.3713 & 0.4342 & 0.6908 & -1.3572 \\
Qwen2.5-3B & 1.0 & 0.1994 & 0.5532 & 0.6910 & -1.5967 \\
Qwen2.5-3B & 1.2 & 0.2973 & 0.4900 & 0.6973 & -1.5291 \\
Qwen2.5-3B & 1.5 & 0.3625 & 0.4409 & 0.6916 & -1.3713 \\

\midrule
\multicolumn{6}{c}{\textbf{Qwen2.5-3B-Instruct}}\\
\midrule
Qwen2.5-3B-Instruct & 0.0 & 0.5403 & 0.2601 & 0.5662 & -0.7109 \\
Qwen2.5-3B-Instruct & 0.1 & 0.5791 & 0.2653 & 0.6304 & -0.6513 \\
Qwen2.5-3B-Instruct & 0.2 & 0.4212 & 0.4192 & 0.7240 & -1.0595 \\
Qwen2.5-3B-Instruct & 0.3 & 0.4146 & 0.4439 & 0.7585 & -1.1307 \\
Qwen2.5-3B-Instruct & 0.4 & 0.2279 & 0.6206 & 0.8039 & -1.6749 \\
Qwen2.5-3B-Instruct & 0.7 & 0.1632 & 0.6625 & 0.7920 & -1.6412 \\
Qwen2.5-3B-Instruct & 1.0 & 0.1166 & 0.6825 & 0.7727 & -1.6085 \\
Qwen2.5-3B-Instruct & 1.2 & 0.2055 & 0.6267 & 0.7889 & -1.6878 \\
Qwen2.5-3B-Instruct & 1.5 & 0.1799 & 0.6409 & 0.7816 & -1.6759 \\

\midrule
\multicolumn{6}{c}{\textbf{TinyLlama-1.1B-Chat-v1.0}}\\
\midrule
TinyLlama-1.1B-Chat-v1.0 & 0.0 & 0.8189 & 0.0885 & 0.4888 & -0.2086 \\
TinyLlama-1.1B-Chat-v1.0 & 0.1 & 0.8511 & 0.0727 & 0.4877 & -0.1696 \\
TinyLlama-1.1B-Chat-v1.0 & 0.2 & 0.9012 & 0.0606 & 0.6130 & -0.1142 \\
TinyLlama-1.1B-Chat-v1.0 & 0.3 & 0.8260 & 0.1476 & 0.8493 & -0.2139 \\
TinyLlama-1.1B-Chat-v1.0 & 0.4 & 0.5661 & 0.4495 & 1.0367 & -0.6482 \\
TinyLlama-1.1B-Chat-v1.0 & 0.7 & 0.2042 & 0.9053 & 1.1380 & -1.7195 \\
TinyLlama-1.1B-Chat-v1.0 & 1.0 & 0.1416 & 0.9795 & 1.1409 & -1.9989 \\
TinyLlama-1.1B-Chat-v1.0 & 1.2 & 0.1449 & 0.9785 & 1.1444 & -1.9792 \\
TinyLlama-1.1B-Chat-v1.0 & 1.5 & 0.2362 & 0.8647 & 1.1317 & -1.5835 \\

\midrule
\multicolumn{6}{c}{\textbf{phi-2}}\\
\midrule
phi-2 & 0.0 & 0.8734 & 0.0785 & 0.6181 & -0.1391 \\
phi-2 & 0.1 & 0.8957 & 0.0675 & 0.6494 & -0.1106 \\
phi-2 & 0.2 & 0.8879 & 0.0920 & 0.8212 & -0.1201 \\
phi-2 & 0.3 & 0.5702 & 0.4483 & 1.0425 & -0.5747 \\
phi-2 & 0.4 & 0.1912 & 0.9145 & 1.1312 & -1.7366 \\
phi-2 & 0.7 & 0.0629 & 1.0733 & 1.1453 & -2.1560 \\
phi-2 & 1.0 & 0.1034 & 1.0362 & 1.1554 & -2.0097 \\
phi-2 & 1.2 & 0.1383 & 0.9918 & 1.1513 & -1.8866 \\
phi-2 & 1.5 & 0.1619 & 0.9653 & 1.1515 & -1.8007 \\

\bottomrule
\end{longtable}

\end{document}